\documentclass[runningheads]{llncs}
\usepackage[T1]{fontenc}
\usepackage{amsmath}
\usepackage{graphicx}
\usepackage{multirow}
\usepackage{booktabs}   
\usepackage{colortbl}   
\usepackage[table]{xcolor} 
\usepackage{graphicx}   
\usepackage{appendix}
\usepackage{tabularx} 
\usepackage{url}
\usepackage{hyperref}

\usepackage{algorithm}
\usepackage{algpseudocode}
\usepackage{booktabs} 

\begin{document}
\title{Satisfiability Modulo Theory Meets Inductive Logic Programming}
%
%
\author{Nijesh Upreti, Vaishak Belle}
\authorrunning{N. Upreti and V. Belle}
%
\institute{The University of Edinburgh, 10 Crichton Street, Edinburgh EH8 9AB, UK}
\maketitle              
%


\begin{abstract}
Inductive Logic Programming (ILP) provides interpretable rule learning in relational domains, yet remains limited in its ability to induce and reason with numerical constraints. Classical ILP systems operate over discrete predicates and typically rely on discretisation or hand-crafted numerical predicates, making it difficult to infer thresholds or arithmetic relations that must hold jointly across examples. Recent work has begun to address these limitations through tighter integrations of ILP with Satisfiability Modulo Theories (SMT) or specialised numerical inference mechanisms. In this paper we investigate a modular alternative that couples the ILP system PyGol with the SMT solver Z3. Candidate clauses proposed by PyGol are interpreted as quantifier-free formulas over background theories such as linear or nonlinear real arithmetic, allowing numerical parameters to be instantiated and verified by the SMT solver while preserving ILP’s declarative relational bias. This supports the induction of hybrid rules that combine symbolic predicates with learned numerical constraints, including thresholds, intervals, and multi-literal arithmetic relations. We formalise this SMT--ILP setting and evaluate it on a suite of synthetic datasets designed to probe linear, relational, nonlinear, and multi-hop reasoning. The results illustrate how a modular SMT--ILP architecture can extend the expressivity of symbolic rule learning, complementing prior numerical ILP approaches while providing a flexible basis for future extensions toward richer theory-aware induction.
\end{abstract}


\section{Introduction}

Inductive Logic Programming (ILP) provides an expressive and interpretable 
framework for learning symbolic rules from relational data 
\cite{muggleton1991inductive,cropper2022inductive}. 
By representing hypotheses in first-order logic, ILP supports strong 
generalisation from limited supervision and allows background knowledge to be 
integrated declaratively into the learning process 
\cite{muggleton2012ilp,evans2018learning,varghese2022efficient}. 
These properties make ILP attractive for domains where relational structure, 
semantic transparency, and explainability are essential.

Classical ILP systems based on inverse entailment 
\cite{muggleton1995inverse} and bottom-clause--guided refinement 
\cite{cropper2022inductive} have been highly successful in symbolic domains. 
However, many real-world problems involve hybrid relational–numerical structures, where relational dependencies interact with numerical quantities, thresholds, and geometric or physical constraints—settings that classical ILP systems struggle to model \cite{speichert2019learning,molina2018mixed}. 
Extending ILP to such hybrid symbolic--numerical settings has proved challenging. 
Traditional approaches rely on discretisation, handcrafted numerical predicates, 
or domain-specific encodings \cite{martin1997learning,kolb2018learning}, 
which often weaken declarative semantics and generalise poorly 
\cite{morel2023inductive}. Early systems such as Aleph explore limited forms 
of numerical reasoning (e.g.\ lazy evaluation), and more recent systems such as 
NumLog \cite{cyrus2025inductive} introduce interval-valued predicates to 
avoid premature threshold selection. These efforts highlight the importance of 
numerical reasoning in ILP but remain restricted in the scope of arithmetic 
relations they can express.

A complementary direction integrates ILP with 
Satisfiability Modulo Theories (SMT). SMT solvers extend the satisfiability problem (SAT) with background 
theories such as linear and nonlinear arithmetic, difference logic, and arrays 
\cite{barrett2009satisfiability,de2008z3,gao2013dreal}, enabling principled 
reasoning over continuous quantities. Recent work such as NUMSYNTH 
\cite{hocquette2023NUMSYNTH} demonstrates that SMT can be used to infer 
numerical constants and arithmetic relationships when combined with ILP-style 
structural search. These developments illustrate the promise of hybrid 
symbolic--numerical learning, but they also raise foundational questions 
regarding how ILP clauses should interface with theory solvers, how feedback 
from SMT should guide hypothesis refinement, and how to balance numerical 
expressivity with interpretability.

In this paper, we investigate these questions through a 
modular SMT--ILP architecture that pairs PyGol \cite{varghese2022efficient} with Z3 \cite{de2008z3}. Rather than tightly integrating the two components, we adopt a 
loosely coupled design: PyGol proposes candidate rule structures, and 
Z3 instantiates and verifies their numerical components. This preserves ILP’s 
declarative workflow while allowing numerical reasoning to be handled entirely 
by the external solver. The modularity of the approach makes it possible to 
inspect, modify, or replace each component independently and to explore how 
different forms of symbolic--numerical interaction affect learning behaviour.

This perspective allows us to study four fundamental issues in hybrid ILP:
\begin{enumerate}
    \item How can symbolic clauses can be compiled systematically into SMT formulas?
    \item How should solver feedback (e.g.\ satisfiable models, unsatisfiable cores) inform pruning and bias selection?
    \item How do linear, relational, and nonlinear numerical relations affect interpretability and rule structure?
    \item How does expressivity and computational tractability interact in 
    modular SMT--ILP systems?
\end{enumerate}

Our goal is not to propose a new end-to-end ILP system or to surpass the 
specialised numerical optimisation capabilities of systems such as NUMSYNTH. 
Instead, we provide a flexible experimental framework for examining 
the interface between symbolic rule induction and theory-aware numerical 
reasoning. Throughout the paper, we use the term SMT--ILP to denote such 
modular architectures in which an ILP engine proposes symbolic rule structures 
and an SMT solver instantiates and verifies their numerical components under 
appropriate background theories.

\section{Motivation}

Our work began with the observation that classical ILP systems based on
inverse entailment, including Progol, Aleph, and other recent variants, rely on
bottom-clause construction to generate the most specific hypothesis for a
single positive example \cite{muggleton1995inverse,srinivasan1999numerical}. While effective for symbolic reasoning, this mechanism 
imposes an overlooked restriction: numerical constants and 
numerical relations can only be drawn from the literals present in that example. As a consequence, bottom-clause-based ILP systems can neither (i) 
synthesise new numerical thresholds, nor (ii) infer arithmetic relationships 
(e.g.\ inequalities, linear separators) that require joint analysis of 
multiple examples. These limitations make it fundamentally difficult 
for traditional ILP to learn concepts that depend on global numerical structure.

PyGol, a recent ILP system built on Meta Inverse Entailment (MIE), improves 
classical ILP by introducing the bottom clause of relevant literals 
and an automatically generated meta-theory that defines a 
double-bounded hypothesis space across all examples \cite{varghese2022efficient}. 
PyGol thereby addresses several shortcomings of Progol’s search and eliminates the need for hand-crafted mode declarations. However, PyGol remains a 
purely symbolic learner; although it can refer to numerical predicates 
present in the background knowledge, it cannot invent thresholds, 
coefficients, or inequalities from data. Any numerical literal that appears in 
a learned rule must originate from an example’s bottom clause. Thus, PyGol can 
discover structural regularities, but it cannot optimise numerical parameters or 
express relations such as 
\[
x < y,\qquad a x + b y \leq c,\qquad x^2 + y^2 < r^2,\qquad xy < c,
\]
even in cases where these constraints define the true decision boundary. Learning such hypotheses requires discovering numerical patterns that only emerge across many examples, a capability that bottom-clause ILP was never designed to support. These limitations reveal a more fundamental gap: classical ILP systems provide relational abstraction, but cannot derive the numerical structure that many concepts depend on. This observation motivates the integration of PyGol with SMT solvers. 

SMT solvers extend propositional SAT solving with 
background theories, including linear and nonlinear real arithmetic, that allow joint 
constraint solving and numerical optimisation \cite{barrett2009satisfiability,de2008z3,gao2013dreal}. Unlike ILP engines, SMT solvers 
are designed to (i) introduce and optimise free numerical parameters, 
(ii) enforce theory consistency across all examples simultaneously, and 
(iii) validate or refute candidate hypotheses using model-theoretic reasoning.

\section{Methodology}

We adopt a modular view of SMT-guided Inductive Logic Programming in which
symbolic structure learning and numerical reasoning are kept explicitly
separate. This design allows us to examine, in a controlled manner, how ILP and SMT interact, which symbolic–numerical tasks benefit from
theory reasoning, and where the limits of such hybrid approaches lie.
Rather than engineering a tightly integrated solver, we treat an ILP engine
(PyGol) and an SMT solver (Z3) as cooperating components that exchange
partial hypotheses and theory constraints: PyGol performs relational
structure learning, predicate invention, and symbolic abstraction, while Z3
instantiates numerical parameters and verifies arithmetic constraints. This
section formalises the resulting SMT--ILP setting and describes the learning
procedure.

\subsection{Problem Formulation}

An SMT--ILP problem instance is given by background knowledge $B$, sets of positive and negative
examples $(E^+,E^-)$, and a hypothesis language $\mathcal{L}_H$ defining the allowable rule
structures. Background knowledge contains both symbolic relations (e.g.\ adjacency, object types)
and interpreted functions from a background theory $T$ such as linear real arithmetic, difference
logic, or nonlinear arithmetic. Examples are treated as partial structures: they may consist
of ground atoms, numerical measurements, or theory constraints.

A hypothesis consists of a clause
\[
h(x) \;:=\; \phi_1(x) \land \cdots \land \phi_k(x),
\]
where each literal $\phi_i$ may be a symbolic predicate, a comparison between
variables (e.g.\ $x < y$), or a theory-specific constraint of the form
$t_1 \,\theta\, t_2$ with $\theta \in \{<,\le,=,\ge,>\}$. Numerical literals
contain symbolic parameters (e.g.\ thresholds, coefficients) that must be
instantiated by the SMT solver. We say that a hypothesis $h$ is
acceptable if, together with the background knowledge $B$, it covers
all positive examples and rejects all negative ones. Rather than testing
logical entailment directly, we express this as a pair of satisfiability
conditions:
\[
\begin{aligned}
& B \land h \land \neg e^+ \;\text{is UNSAT for all } e^+ \in E^+ ,\\
& B \land h \land e^-      \;\text{is UNSAT for all } e^- \in E^- ,
\end{aligned}
\]
where satisfiability is checked with respect to theory~$T$. In other words,
no model of $B \land h$ may falsify a positive example or satisfy a negative
one. This shifts ILP from symbolic unification over ground atoms to reasoning
over mixed symbolic--numerical constraints.

\subsection{Modular SMT--ILP Learning}

Learning proceeds by decoupling the generation of rule \emph{structure} from the instantiation of
numerical parameters.

\paragraph{(1) Structural hypothesis generation (PyGol).}
PyGol generates candidate clauses using Meta-Interpretive refinement. Literal schemas define the
search space, and, when enabled, predicate invention introduces auxiliary predicates that capture
multi-step relational patterns. Numerical literals produced during this phase contain uninstantiated
numerical parameters.

\paragraph{(2) Parameter instantiation and constraint solving (Z3).}
For each candidate structure, Z3 constructs a MaxSMT problem whose free variables represent numerical
parameters (e.g.\ interval bounds, linear coefficients, distance thresholds). Positive examples
impose hard constraints ensuring coverage, whereas negative examples impose soft constraints that
reward correct rejection. Solving the MaxSMT instance yields parameter values that best separate the
examples under theory~$T$.

\paragraph{(3) Verification and pruning.}
Instantiated rules are checked for theory satisfiability, and incompatible or overly specific rules
are removed. Rules are then ranked using a score based on coverage and precision.

\paragraph{(4) Iterative refinement.}
Accepted rules are added to the background knowledge. PyGol may then reuse them in subsequent
structural search, allowing the system to construct increasingly expressive relational or hybrid
constraints across iterations.

\subsection{General Properties of the Framework}
This modular design yields three key advantages. First, it maintains a clear separation between structure and numerics: PyGol focuses on discovering relational patterns without committing to specific numerical values, which are subsequently determined by Z3. Second, the approach supports theory-aware learning, since Z3 reasons directly within background theories. As a result, hypotheses may include linear, interval, quadratic, multiplicative, or even trigonometric constraints without requiring changes to the ILP search procedure. Third, the framework offers a degree of model-theoretic robustness: candidate hypotheses are validated against partial numerical information provided by examples, allowing the system to learn effectively even when measurements are incomplete. The next section examines how this general learning procedure specialises across datasets and how PyGol and Z3 interact in practice in linear, relational, and nonlinear domains.

\section{Experimental Setup}

Our experiments evaluate how a modular SMT--ILP architecture behaves across problems requiring symbolic, numerical, nonlinear, and relational reasoning. The system consists of PyGol that performs structure learning via Meta-Interpretive search, and Z3 that supplies theory reasoning, constraint optimisation, and numerical parameter inference. Importantly, we do not engineer a unified solver: the system remains loosely coupled, with PyGol proposing candidate rule structures and Z3 validating, refining, or rejecting these structures through theory-level reasoning. This design allows us to analyse the role of SMT in different ILP regimes without conflating structural and numerical search.

\subsection{SMT--ILP Integration Pipeline}

We model learning as an iterative process in which ILP proposes structural hypotheses and an SMT
solver instantiates and verifies their numerical components. Starting from initial background
knowledge $B_0$, each iteration $t = 1,2,\ldots$ follows the same ILP--SMT interaction pattern:
PyGol first generates candidate clause structures (optionally inventing predicates), after which
numerical literals within these clauses are treated as symbolic parameters. Z3 then solves a
dataset-specific MaxSMT problem to instantiate these parameters, checks rule satisfiability under
the background theory, and returns only numerically feasible clauses. Validated rules are scored,
selected, and incorporated into the background knowledge before the next iteration.

The full procedure is summarised in Procedure~\ref{fig:smt-ilp-loop}. This modular design separates
structural and numerical reasoning: PyGol provides relational and logical generalisation, while Z3
performs theory-guided optimisation and verification across increasingly expressive numerical
domains.

\subsection{Dataset-Specific Learning Behaviours}

Although the learning loop is uniform, the interaction pattern between PyGol and Z3 differs markedly across datasets, each of which stresses a different dimension of symbolic--numerical integration.

\paragraph{Geometry0: Basic Linear Constraints.}
These two tasks (interval, halfplane) isolate arithmetic learning. PyGol proposes simple one-literal or two-literal clauses; Z3 learns either interval bounds or coefficients of 2D linear inequalities. Z3 solves small MaxSMT instances with hard constraints on positives and soft constraints on negatives.

\paragraph{Geometry1: 3D and Conjunctive Constraints.}
Here arithmetic learning extends to three variables. Z3 infers expressions of the form $a x + b y + c z \le d$, and many tasks require conjunctions of independently learned constraints (e.g.\ a halfspace and an interval). PyGol identifies the structural decomposition, and Z3 instantiates each component numerically.

\paragraph{Geometry2: Relational Spatial Constraints.}
These ten tasks (e.g.\ left\_of, inside) require comparing multiple head arguments. The difficulty is relational: NUMSYNTH-style constant learning is insufficient because many rules involve comparisons between variables (e.g.\ $X_1 < X_2$). PyGol must discover these relational patterns; Z3 refines thresholds where applicable (e.g.\ distance or bounding-box limits). This dataset stresses ILP's relational expressivity and SMT's ability to validate numerical subcomponents of relational rules.

\paragraph{Geometry3: Nonlinear and Disjunctive Constraints.}
These fourteen tasks require quadratic (circles, parabolas), multiplicative (hyperbolic regions), trigonometric (sinusoids), or disjunctive (unions of regions) constraints. PyGol proposes decompositions of complex regions into clauses; Z3 instantiates nonlinear arithmetic parameters and verifies satisfiability. SMT reasoning is indispensable here: ILP alone cannot express or fit such numerical relations.

\paragraph{InfluencePropagation (IP): Multi-Hop Relational Reasoning with Predicate Invention.}
These five tasks differ fundamentally from the Geometry datasets. Although influence computations involve nonlinear arithmetic, the primary challenge is discovering multi-hop relational patterns such as $A \rightarrow B \rightarrow C \rightarrow A$. Such patterns cannot be represented at bounded clause depth without predicate invention. PyGol invents auxiliary predicates capturing intermediate relational abstractions; Z3 optimises numerical thresholds associated with influence-based predicates. Thus Geometry datasets stress SMT's numerical expressivity, whereas IP stresses ILP's structural expressivity and the necessity of predicate invention.

\subsection{Training Procedure and Hyperparameters}

All datasets use fixed random seeds and a 70/30 train--test split. PyGol is configured with
dataset-dependent literal budgets reflecting the structural complexity of each benchmark.
Geometry0 uses up to three literals per clause, Geometry1 and Geometry3 use up to six
literals, and Geometry2 uses up to five. For the InfluencePropagation (IP) tasks, we fix the literal budget to four in all configurations, including ablations without predicate invention, since the
target multi-hop patterns require at least three body literals and cannot be expressed under a two-literal budget. Timeouts are likewise dataset-specific: Geometry0 uses a 45\,s timeout, Geometry1 a 120\,s timeout, Geometry2 a 30\,s timeout, Geometry3 a 60\,s timeout, and IP tasks use timeouts ranging from 30--180\,s depending on task complexity. On the SMT side, Z3 performs MaxSMT optimisation using coefficient bounds in $[-100,100]$. Nonlinear arithmetic is enabled for
Geometry3, and threshold optimisation is applied universally to clauses involving distances,
spatial relations, or influence computations.

\subsection{Evaluation Protocol and Reproducibility}

Our evaluation focuses on the interaction between ILP structure learning and SMT-driven
numerical reasoning. We report test accuracy under standard ILP entailment and average runtime
per iteration. For the IP tasks, we additionally compare performance with and without
predicate invention to isolate its contribution to multi-hop relational reasoning. Across
all datasets, we also provide qualitative analyses of representative learned rules to
illustrate differences in numerical, relational, and nonlinear behaviour.

To assess robustness under dataset-level randomness, Geometry0 and Geometry1 are evaluated over ten independent trials, whereas Geometry2, Geometry3, and the IP benchmarks are evaluated over five trials due to their greater computational cost. Reported results reflect means and standard deviations across trials.

\section{Results}

We evaluate the behavior of our modular SMT--ILP architecture across five
benchmark families of increasing symbolic and numerical complexity\footnote{Code available at \url{https://github.com/nuuoe/SMT-ILP}.}. Our analysis
addresses two questions: (i) how PyGol--Z3 performs across linear, relational,
nonlinear, and multi-hop domains, and (ii) how this modular design compares
with specialised numerical ILP systems such as NUMSYNTH. Because the current
NUMSYNTH implementation is designed for hypotheses with single-argument heads
and constant-based numerical comparisons, direct comparison is meaningful only on
the purely numerical benchmarks (Geometry0--1). For the remaining datasets—
which require variable--variable comparisons, nonlinear theories, predicate
invention, or disjunctive structure—the NUMSYNTH language does not currently
apply, and we therefore report only PyGol--Z3 performance. We emphasise that these observations concern the current implementation of NUMSYNTH rather than a fundamental limitation of the approach; extending its numerical abstraction mechanisms to richer rule forms remains a promising direction for future work. We also do not compare against NumLog, which focuses on interval-based numeric reasoning and does not operate in the hybrid relational–numeric settings required by the Geometry and IP benchmarks. Exploring such comparisons is left for future work.

\subsection{Comparison with NUMSYNTH on Pure Numerical Benchmarks}

Geometry0 and Geometry1 contain linear numerical classification tasks closely
aligned with the design assumptions of the current NUMSYNTH system. 
Table~\ref{tab:geom01_comp} summarises performance on Geometry0. As expected, 
NUMSYNTH achieves perfect accuracy with minimal runtime, reflecting its 
specialised optimisation pipeline and efficient constant-learning mechanism. 
PyGol--Z3 also attains high accuracy, though with slightly lower performance 
on the interval task and higher runtime due to structural clause enumeration and 
MaxSMT optimisation.

As shown in Table~\ref{tab:geom01_comp} comparison for Geometry1, NUMSYNTH again performs strongly on tasks that fall within its supported 
hypothesis language (3D halfspaces and multi-halfspace conjunctions). 
However, in tasks requiring higher-dimensional intervals or mixed conjunctive 
constraints, the current NUMSYNTH implementation encounters substantial 
slowdowns or timeouts. PyGol--Z3, by contrast, solves all Geometry1 tasks, 
albeit at higher computational cost and with varying accuracy depending on 
numerical complexity. 

The Geometry0 and Geometry1 comparison highlight a trade-off. NUMSYNTH offers
state-of-the-art efficiency and accuracy when tasks match the numerical forms
expressible in its present hypothesis language. PyGol--Z3, in contrast, 
separates symbolic structure learning from numerical parameter optimisation, 
resulting in slower runtimes but a substantially broader expressive range. 
This modularity becomes essential once tasks require relational structure, 
variable--variable comparisons, nonlinear arithmetic, or predicate invention—
settings where NUMSYNTH's current design does not yet operate.

This complementary relationship suggests potential avenues for future work: 
NUMSYNTH’s efficient numerical abstraction and optimisation mechanisms might be 
extended to accommodate richer relational or theory-aware literals, or 
combined with PyGol’s flexible structural search. Such hybrids could unify the 
strengths of both approaches while maintaining interpretability.

\begin{table}[t]
\centering
\begin{tabular}{l|c|c|c|c}
\hline
\textbf{Task} & \textbf{ NS Accuracy } & \textbf{ PZ Accuracy } &
\textbf{ NS Time } & \textbf{ PZ Time } \\
\hline
\multicolumn{5}{c}{\textbf{Geometry0}} \\
\hline
Halfplane            & $100 \pm 0$ & $94 \pm 1$ & $5 \pm 1$   & $4 \pm 0$ \\
Interval             & $100 \pm 0$ & $91 \pm 1$ & $2 \pm 0$   & $2 \pm 0$ \\
\hline
\multicolumn{5}{c}{\textbf{Geometry1}} \\
\hline
3D Halfplane         & $100 \pm 0$ & $94 \pm 1$ & $365 \pm 34$ & $37 \pm 1$ \\
Conjunction          & Timeout     & $78 \pm 6$ & Timeout      & $68 \pm 3$ \\
3D Interval          & Timeout     & $81 \pm 4$ & Timeout      & $58 \pm 2$ \\
Multiple Halfplanes  & $100 \pm 0$ & $84 \pm 4$ & $4 \pm 1$    & $18 \pm 1$ \\
\hline
\end{tabular}
\vspace{0.4em}
\caption{\textbf{Comparison of NUMSYNTH (NS) and PyGol--Z3 (PZ) on Geometry0--1 (10 trials).}  
NUMSYNTH performs well on simple linear tasks but times out on conjunctive and mixed 3D constraints,  
while PyGol--Z3 solves all tasks with reasonable accuracy.}
\label{tab:geom01_comp}
\vspace{-2em}
\end{table}

\subsection{Relational Constraints: Geometry2}

Geometry2 consists of ten tasks requiring relational comparisons between
multiple head arguments, such as spatial ordering, containment,
alignment, and distance relations. These concepts cannot be expressed in ILP
systems that restrict numerical literals to variable--constant comparisons, nor
in numerical ILP systems such as NUMSYNTH that prohibit relations of the form
$X_1 < X_2$.

A representative example is the \texttt{left\_of} task, where the target
concept is:
\[
\texttt{left\_of}(P,Q) \leftarrow x(P) < x(Q).
\]
Here PyGol must propose a relational rule schema linking two points, while Z3
instantiates the threshold-free inequality $x(P) < x(Q)$. Another example is
\texttt{inside}, where the target region is specified by bounding-box
constraints:
\[
\texttt{inside}(P,R) \leftarrow x_{\min}(R) \le x(P) \le x_{\max}(R),
\,
y_{\min}(R) \le y(P) \le y_{\max}(R).
\]
PyGol discovers the relational decomposition (linking $P$ to region $R$),
while Z3 jointly optimises the interval constraints.

More complex tasks such as \texttt{between} require relational patterns that 
cannot be expressed without variable--variable comparisons and geometric 
constraints. The \texttt{between} task requires point $P$ to be both collinear 
with and between points $A$ and $B$:

\[
\texttt{between}(P,A,B) \leftarrow 
\text{collinear}(A,P,B) \;\land\; \text{between}_g(A,P,B).
\]

The collinearity predicate is enforced by the cross-product condition
\[
(x(P)-x(A))(y(B)-y(A))
\;-\;
(y(P)-y(A))(x(B)-x(A)) \approx 0,
\]
and the geometric betweenness predicate $\text{between}_g(A,P,B)$ is enforced by the dot-product condition
\[
(x(P)-x(A))(x(B)-x(P))
\;+\;
(y(P)-y(A))(y(B)-y(P)) \ge 0.
\]

Here PyGol searches for a three-argument relational schema, and Z3 verifies 
the collinearity and betweenness constraints across all examples.

Table~\ref{tab:geom2_results} shows that PyGol--Z3 achieves high accuracy on
most tasks, solving even structurally complex ones such as \texttt{aligned},
\texttt{overlapping}, or \texttt{surrounds}. Runtimes vary, reflecting the
cost of discovering appropriate relational schemas and solving multiple
threshold constraints jointly. Geometry2 therefore demonstrates the first
regime in which the modular design is essential: the ILP component handles
relational generalisation, while the SMT solver instantiates numerical
constraints that no ILP system could infer directly.

\begin{table}[t]
\centering
\begin{tabular}{l|c|c}
\hline
\textbf{Task} & \textbf{ Accuracy (\%) } & \textbf{ Time (s) } \\
\hline
left\_of        & $94 \pm 3$  & $7 \pm 0$   \\
closer\_than    & $91 \pm 2$  & $34 \pm 2$  \\
touching        & $94 \pm 3$  & $8 \pm 0$   \\
inside          & $94 \pm 0$  & $131 \pm 4$ \\
overlapping     & $98 \pm 2$  & $361 \pm 49$ \\
between         & $76 \pm 5$  & $31 \pm 1$  \\
adjacent        & $73 \pm 0$  & $698 \pm 69$ \\
aligned         & $82 \pm 13$ & $785 \pm 58$ \\
surrounds       & $99 \pm 1$  & $159 \pm 24$ \\
near\_corner    & $100 \pm 0$ & $59 \pm 1$  \\
\hline
\end{tabular}
\vspace{0.5em}
\caption{\textbf{Geometry2 performance (5 trials).}
Predictive accuracy (\%) and runtime (s) for relational spatial tasks requiring
variable–variable comparisons.}
\label{tab:geom2_results}
\vspace{-2em}
\end{table}

\subsection{Nonlinear and Disjunctive Constraints: Geometry3}

Geometry3 introduces nonlinear, multiplicative, trigonometric, and
disjunctive decision boundaries. These tasks cannot be represented in classical
ILP, nor in numerical ILP systems restricted to linear templates. They therefore
serve as a clean test of whether PyGol--Z3 can discover symbolic structure
sufficient to support nonlinear numerical optimisation.

As a simple example, the \texttt{in\_circle} task requires expressing the
constraint:

\[
x(P)^2 + y(P)^2 \le r^2,
\]

which cannot be manufactured by ILP alone because it introduces new nonlinear
terms and a free parameter $r$. PyGol proposes the relational skeleton
linking point $P$ to its coordinates, and Z3 fits the radius.

More complex examples include disjunctive regions such as
\texttt{circle\_or\_box}:

\[
\texttt{target}(P) \leftarrow 
(x(P)^2 + y(P)^2 \le r^2)
\;\lor\;
(|x(P)| \le s \land |y(P)| \le s),
\]

where PyGol must decompose the concept into multiple alternative clauses,
while Z3 fits separate parameter sets for each branch.

The \texttt{crescent} and \texttt{donut} tasks highlight the role of symbolic
decomposition. For a donut region:

\[
r_{\min}^2 \le x(P)^2 + y(P)^2 \le r_{\max}^2,
\]

PyGol learns two constraints (inner and outer) while Z3 fits both radii.
Similarly, the \texttt{sinusoidal} task requires fitting parameters in

\[
y(P) \ge \sin(\omega x(P)) + \phi,
\]

where $\omega$ is the frequency scale and $\phi$ is the vertical offset, which
relies entirely on Z3's ability to solve nonlinear arithmetic.

Table~\ref{tab:geom3_results} shows that PyGol--Z3 achieves high accuracy
across the full suite, with variations in runtime reflecting the complexity of
nonlinear MaxSMT optimisation. Geometry3 provides clear evidence that
modular ILP--SMT reasoning substantially extends the expressive power of ILP,
enabling learning in domains that are inherently inaccessible to symbolic
search alone.

\begin{table}[t]
\centering
\begin{tabular}{l|cc}
\toprule
\textbf{Task} & \textbf{Accuracy (\%)} & \textbf{Time (s)} \\
\midrule
in\_circle        & $91 \pm 4$   & $67 \pm 1$ \\
in\_ellipse       & $89 \pm 5$   & $786 \pm 17$ \\
hyperbola\_side   & $91 \pm 3$   & $71 \pm 1$ \\
xy\_less\_than    & $91 \pm 2$   & $20 \pm 1$ \\
quad\_strip       & $94 \pm 3$   & $21 \pm 1$ \\
union\_halfplanes & $94 \pm 3$   & $49 \pm 1$ \\
circle\_or\_box   & $100 \pm 0$  & $341 \pm 5$ \\
piecewise         & $81 \pm 3$   & $60 \pm 3$ \\
fallback\_region  & $94 \pm 0$   & $450 \pm 7$ \\
donut             & $100 \pm 0$  & $98 \pm 2$ \\
lshape            & $89 \pm 5$   & $55 \pm 1$ \\
above\_parabola   & $98 \pm 2$   & $149 \pm 2$ \\
sinusoidal        & $98 \pm 2$   & $137 \pm 3$ \\
crescent          & $93 \pm 4$   & $1243 \pm 13$ \\
\bottomrule
\end{tabular}
\vspace{0.4em}
\caption{\textbf{Geometry3 performance (5 trials).}
Predictive accuracy (\%) and runtime (s) for tasks requiring nonlinear, multiplicative, trigonometric, or disjunctive constraints.}

\label{tab:geom3_results}
\vspace{-2em}
\end{table}

\subsection{Multi-Hop Relational Reasoning with Predicate Invention: IP Tasks}

The InfluencePropagation (IP) benchmark tests two capabilities that classical
ILP systems struggle with simultaneously: (i) multi-hop relational
dependencies and (ii) numerical influence thresholds. The true concepts
often involve multi-step propagation patterns such as
\[
\texttt{active}(A) \leftarrow 
\texttt{edge}(A,B),\ \texttt{edge}(B,C),\ \texttt{score}(C) > \tau,
\]
which cannot be represented without introducing auxiliary predicates
capturing intermediate relational structure.

For example, in \texttt{ip3\_threshold}, the correct hypothesis involves
a triangle pattern (A→B→C→A) with an influence threshold:

\[
\begin{aligned}
\texttt{reach2}(A,C) &\leftarrow \texttt{propagates}(A,B), \texttt{propagates}(B,C),\\
\texttt{active}(A) &\leftarrow \texttt{reach2}(A,C), \texttt{propagates}(C,A),\\
                     &\quad \texttt{max\_influence}(A) > \tau.
\end{aligned}
\]

PyGol is responsible for inventing the predicate \texttt{reach2} to capture
the two-hop chain; Z3 is responsible for fitting the influence threshold $\tau$.

\begin{table}[t]
\centering
\begin{tabular}{l|c|c|c|c}
\hline
\textbf{Task} &
\textbf{ No PI } &
\textbf{ PI Only } &
\textbf{ PI+Z3 } &
\textbf{ Time (s) } \\
\hline
ip1\_active      
& $100 \pm 0$
& $100 \pm 0$
& $100 \pm 0$
& $89 \pm 17$ \\
ip2\_active      
& N/A
& $78 \pm 4$
& $85 \pm 4$
& $153 \pm 12$ \\
ip3\_active      
& N/A
& $67 \pm 7$
& $76 \pm 6$
& $159 \pm 11$ \\
ip3\_threshold   
& N/A
& $56 \pm 3$
& $79 \pm 4$
& $129 \pm 23$ \\
ip4\_high\_score 
& N/A
& $48 \pm 3$
& $83 \pm 6$
& $164 \pm 32$ \\
\hline
\end{tabular}
\vspace{0.4em}
\caption{\textbf{IP benchmark performance (5 trials).}
Accuracies for five tasks under three configurations—no predicate invention (No PI),
predicate invention only (PI only), and predicate invention with SMT optimisation (PI+Z3). Reported runtimes correspond to PI+Z3.}
\label{tab:ip_results}
\vspace{-2em}
\end{table}

The results in Table~\ref{tab:ip_results} reveal three key findings:

\begin{enumerate}
    \item \textit{Without predicate invention}, the necessary relational
    structure is simply inexpressible, leading to universal failure beyond the
    simplest task.

    \item \textit{Predicate invention alone} recovers the relational structure
    but yields poor accuracy because influence thresholds cannot be tuned
    symbolically.

    \item \textit{The full PyGol--Z3 system} succeeds across all tasks, with
    SMT contributing improvements of up to 35 percentage points on
    threshold-heavy tasks.
\end{enumerate}

Thus the IP tasks illustrate a dual dependence: symbolic invention is required
to express multi-hop structure, while SMT reasoning is required to calibrate
numerical influence. Neither mechanism alone suffices.

Across the five benchmark families, we observe a clear progression in
representational demands.  NUMSYNTH achieves state-of-the-art performance on
purely numerical tasks but is unable to express relational, nonlinear, or
multi-hop dependencies.  PyGol--Z3 is competitive on simple numerical problems
and exhibits increasing relative strength as task structure becomes more
complex.  The modular architecture enables PyGol to discover symbolic
structure, while Z3 fills in numerical details ranging from linear thresholds to
nonlinear parameters.  This division of labour ultimately allows the system to
handle tasks that fall well outside the design envelope of existing ILP or
numerical ILP engines.

\section{Related Work}

Hybrid symbolic–numerical reasoning, understood as learning rules that combine relational symbolic structure with continuous numerical constraints, has been a long-standing challenge in ILP. Early ILP systems were designed for purely symbolic domains 
\cite{muggleton1991inductive,muggleton1994inductive,muggleton1995inverse,Bergadano1995}, and their mechanisms---such as inverse entailment, bottom-clause construction, and refinement-based search \cite{cropper2022inductive}---do not natively support arithmetic constraints or continuous variables. As a result, much of the classical work on numerical ILP relied on discretisation, hand-crafted numerical predicates, or domain-specific encodings.

Methods such as TILDE \cite{blockeel1998top} discretise continuous variables by selecting symbolic thresholds during decision tree induction. While efficient, discretisation often obscures the underlying numerical structure and may fail to generalize when examples are sparse. Similarly, QCBA \cite{kliegr2023qcba} refines association-rule learners for quantitative attributes but still relies on discretised numerical intervals, which can lead to loss of information and unnecessarily complex rules. These limitations highlight the need for ILP systems that handle numerical constraints as arithmetic objects rather than reducing them to symbolic abstractions. Several early systems aimed to incorporate numerical constraints more directly. ICC \cite{martin1997learning} embedded real-valued constraints into ILP clause structures, while the NUM algorithm \cite{anthony1997generating} integrated numerical reasoning into ILP by coupling refinement with Constraint Logic Programming \cite{jaffar1994constraint}. FORS system \cite{karalivc1997first} combined ILP with numerical regression, allowing continuous target functions inside logical structures. Constraint inductive logic programming framework \cite{sebag1996constraint} modelled numerical constraints using difference logic but were limited to generalizing a single positive example. These systems demonstrated the value of numerical reasoning but lacked support for shared numerical variables across multiple literals or joint reasoning over multiple examples.

The lazy evaluation extension \cite{srinivasan1999numerical,srinivasan1997experiments} provided a mechanism for discovering numerical values by delaying evaluation of numerical predicates until partial hypotheses were instantiated on examples. While capable of predicting numerical outputs via custom loss functions \cite{srinivasan1997experiments}, lazy evaluation treats each numerical literal independently. This prevents Aleph from learning hypotheses requiring coordinated numerical constraints across multiple literals (e.g., learning both an upper and a lower bound for the same variable). Aleph also requires users to supply background definitions for numerical computations, does not support predicate invention, and struggles with recursive or textually minimal programs. Recent learners \cite{cropper2021popper,purgal2022lff} enumerate rule templates or constant symbols as unary predicates, preventing learning over infinite numerical domains. MAGICPOPPER \cite{hocquette2023magicpopper} extends lazy evaluation by allowing symbolic variables to represent unknown constants but evaluates hypotheses independently on each example, preventing joint numerical inference across examples. 

Among prior systems, NUMSYNTH is the most directly related to our setting. The NUMSYNTH framework \cite{hocquette2023NUMSYNTH} introduced an ILP system capable of learning numerical constants from infinite domains using SMT. NUMSYNTH decomposes learning into (i) structural search over partial hypotheses using Popper \cite{cropper2021popper} and (ii) SMT-based inference of numerical values across all examples simultaneously. It can learn linear constraints such as intervals, thresholds, and chained numerical relations. However, NUMSYNTH tightly couples ILP search and SMT solving inside a custom engine, making the architecture difficult to generalize or modularize. Its reliance on precise numerical values also creates risks of overfitting in sparse numerical settings.

A related but conceptually distinct line of work is NumLog \cite{cyrus2025inductive}, which induces interval-valued numerical predicates rather than point-valued constants. This improves interpretability and mitigates over-precise thresholding, but NumLog does not incorporate SMT and cannot express richer arithmetic relationships, limiting its applicability to numerical constraints that can be captured purely through interval reasoning.

SMT solvers \cite{de2008z3,gao2013dreal} have also been widely adopted in constraint-acquisition and declarative learning frameworks. Systems such as \cite{kolb2018learning,de2018learning} learn numerical constraints over flat attribute spaces, while satisfiability-guided approaches \cite{morel2023inductive,mocanu2019pac+} infer symbolic–numerical constraints under model-theoretic or PAC semantics. Although effective, these methods lack relational representations and therefore do not address how first-order hypothesis search should interact with theory-level arithmetic. Related developments in program synthesis also make extensive use of SAT/SMT solving: Sketch \cite{solar2013program} completes partial programs via counterexample-guided inductive synthesis using SAT solvers; Reynolds et al.\ present an SMT-based synthesis engine integrated into CVC4 \cite{reynolds2015counterexample} that extracts functions from unsatisfiability proofs and supports syntax-guided constraints; Jha et al.\ \cite{jha2010oracle} combine oracle-guided learning with SMT-based component synthesis for loop-free programs; and Albarghouthi et al.\ \cite{albarghouthi2017constraint} synthesize recursive Datalog programs using SMT solver-driven constraint reasoning. Together, these approaches demonstrate the versatility of SMT in guiding structured search and constructing programs or constraints from partial specifications, offering methodological inspiration for exploring alternative ways ILP and SMT might be combined.

\section{Discussion and Future Directions}

Our study highlights both the opportunities and the challenges of integrating SMT solving into an ILP workflow. In contrast to tightly coupled systems such as NUMSYNTH \cite{hocquette2023NUMSYNTH}—where hypothesis construction and numerical reasoning are interleaved inside a single engine—our modular PyGol--Z3 architecture separates symbolic rule induction from numerical constraint solving. This makes it possible to examine, in a controlled manner, how relational hypotheses should be mapped into theory-level constraints, how solver feedback ought to guide search, and what classes of geometric and arithmetic patterns can be learned reliably from finite examples.

A central challenge revealed by our experiments is the granularity mismatch between symbolic ILP and SMT solvers. ILP explores structural hypotheses over first-order predicates, whereas SMT solvers operate over fully instantiated arithmetic constraints. Bridging these layers requires careful design of bias languages, hypothesis templates, and solver encodings, especially as symbolic–numerical rules grow combinatorially with the number of predicates and theory operators. Prior work on solver-guided ILP such as NUMSYNTH \cite{hocquette2023NUMSYNTH} shows that numerical inference across multiple examples is feasible, but also demonstrates the risks of overfitting when numerical thresholds are treated as precise point values. Our modular approach opens the door to solver-aware pruning, caching of numerical evaluations, and staged refinement, which may significantly improve tractability in future systems.

Although our experiments focus on linear arithmetic, SMT solvers support richer theories—including non-linear arithmetic, bit-vectors, arrays, and algebraic datatypes \cite{de2008z3,gao2013dreal}. These capabilities remain under-explored in ILP. A modular system such as ours provides a natural platform to investigate theory selection and mixed-theory reasoning, enabling future work on non-convex geometric concepts, piecewise-defined relations, and mixed integer–real domains. This extends beyond the numerical fragments targeted by NUMSYNTH and may help bridge ILP with broader SMT-based learning frameworks \cite{morel2023inductive}.

A further challenge concerns predicate invention in numerical domains. Predicate invention is powerful for relational abstraction, but our results suggest that unconstrained invention risks overfitting or producing uninterpretable constructs when numerical parameters are involved. This echoes findings from earlier ILP systems that struggled to coordinate numerical constraints across invented predicates \cite{srinivasan1999numerical}. A promising future direction is to constrain invention using solver-derived signals—for example, unsatisfiable cores, theory lemmas, or recurring arithmetic substructures that appear across multiple candidate hypotheses.

Another fruitful avenue is probabilistic and approximate reasoning. Work on weighted model integration and probabilistic inference over hybrid domains \cite{belle2015probabilistic,chistikov2015approximate} demonstrates that continuous constraints can be handled in soft or approximate forms, offering robustness against noise. Integrating such probabilistic semantics into an ILP–SMT workflow could enable hypothesis scoring under uncertainty and principled treatment of noisy geometric observations.

Our results underscore the importance of adaptive inductive bias. At present, systems like PyGol require manually specified rule templates, constraints, and relevant background predicates. Developing meta-learning or bias-adaptation mechanisms—potentially inspired by declarative bias learning frameworks or constraint-acquisition methods \cite{de2018learning,kolb2018learning}—could allow ILP–SMT systems to tune their hypothesis spaces, solver parameters, or theory fragments automatically. This would transform ILP–SMT from a hand-engineered pipeline into a more autonomous and data-driven reasoning framework.

An additional direction concerns how SMT--ILP could interface with developments in differentiable and neuro-symbolic learning. Differentiable ILP models \cite{shindo21,krishnan21} learn soft logical rules through gradient descent and may provide useful inductive biases or template priors that can guide SMT-based rule search. Neuro-symbolic approaches \cite{speichert2019learning,bueff2024learning} similarly offer mechanisms for learning feature abstractions or latent geometric representations that could be translated into SMT constraints, helping the solver prune or prioritize candidate hypotheses. Likewise, statistical relational learning \cite{kersting2011statistical} introduces probabilistic semantics for first-order structure, suggesting possible extensions of SMT--ILP in which numerical constraints are reasoned about under uncertainty rather than deterministically. Although these paradigms operate with different assumptions and objectives, they offer promising complementary tools for building SMT--ILP systems that are more adaptive, robust, and data-driven.

Overall, the modular ILP–SMT architecture presented here should be regarded as an initial, highly limited exploration rather than a mature system. At present, our framework only supports small, hand-specified hypothesis languages, relies on simple numerical constraints, and does not yet include automated bias adaptation or sophisticated solver–search interaction. These restrictions mean that SMT--ILP is far less capable than fully engineered systems such as NUMSYNTH or more specialized approaches like NumLog \cite{cyrus2025inductive}. Nevertheless, the value of our contribution lies in establishing a minimal but extensible foundation for studying ILP–SMT interaction in a modular way. Future work will focus on expanding expressivity, automating inductive bias selection, enriching numerical abstraction mechanisms, and developing tighter feedback loops between ILP search and SMT reasoning.

\section{Conclusion}
In this paper, we introduced a modular framework that augments ILP with the arithmetic reasoning capabilities of SMT. By allowing candidate rules to be evaluated against background theories such as linear real arithmetic, PyGol--Z3 can express hypotheses that combine relational structure with simple numerical constraints while retaining the declarative semantics and interpretability of classical ILP. Our aim has not been to engineer a full SMT+ILP \cite{belle2020smt+} system, but rather to establish a minimal, transparent substrate on which such systems can be studied. Through a model-theoretic formulation and small illustrative examples, we showed how partial models and solver-based consistency checks can support rule induction in hybrid domains. Although the present framework remains limited in scope and functionality, it provides a basis for future work on richer arithmetic reasoning, solver-guided refinement strategies, automated bias adaptation, and potential integrations with differentiable or probabilistic extensions to ILP.

\bibliographystyle{splncs04}
\bibliography{references}

@article{solar2013program,
  title={{Program Sketching}},
  author={Solar-Lezama, Armando},
  journal={International Journal on Software Tools for Technology Transfer},
  volume={15},
  number={5},
  pages={475--495},
  year={2013},
  publisher={Springer}
}

@inproceedings{jha2010oracle,
  title={{Oracle-guided Component-based Program Synthesis}},
  author={Jha, Susmit and Gulwani, Sumit and Seshia, Sanjit A and Tiwari, Ashish},
  booktitle={Proceedings of the 32nd ACM/IEEE International Conference on Software Engineering-Volume 1},
  pages={215--224},
  year={2010}
}

@inproceedings{albarghouthi2017constraint,
  title={{Constraint-based Synthesis of Datalog Programs}},
  author={Albarghouthi, Aws and Koutris, Paraschos and Naik, Mayur and Smith, Calvin},
  booktitle={International Conference on Principles and Practice of Constraint Programming},
  pages={689--706},
  year={2017},
  organization={Springer}
}

@inproceedings{reynolds2015counterexample,
  title={{Counterexample-guided Quantifier Instantiation for Synthesis in SMT}},
  author={Reynolds, Andrew and Deters, Morgan and Kuncak, Viktor and Tinelli, Cesare and Barrett, Clark},
  booktitle={International Conference on Computer Aided Verification},
  pages={198--216},
  year={2015},
  organization={Springer}
}

@article{purgal2022lff,
  title={{Learning Higher-order Logic Programs from Failures}},
  author={Purga{\l}, Stanis{\l}aw J and Cerna, David M and Kaliszyk, Cezary},
  journal={IJCAI 2022},
  pages={2726--2733},
  year={2022}
}

@article{cropper2021popper,
  title={{Learning Programs by Learning from Failures}},
  author={Cropper, Andrew and Morel, Rolf},
  journal={Machine Learning},
  volume={110},
  number={4},
  pages={801--856},
  year={2021},
  publisher={Springer}
}

@article{hocquette2023magicpopper,
  title={{Learning Programs with Magic Values}},
  author={Hocquette, C{\'e}line and Cropper, Andrew},
  journal={Machine Learning},
  volume={112},
  number={5},
  pages={1551--1595},
  year={2023},
  publisher={Springer}
}

@inproceedings{hocquette2023NUMSYNTH,
  title={{Relational Program Synthesis with Numerical Reasoning}},
  author={Hocquette, C{\'e}line and Cropper, Andrew},
  booktitle={Proceedings of the AAAI Conference on Artificial Intelligence},
  volume={37},
  number={5},
  pages={6425--6433},
  year={2023}
}

@article{srinivasan1997experiments,
  title={{Experiments in Numerical Reasoning with Inductive Logic Programming}},
  author={Srinivasan, Ashwin and Camacho, Rui},
  journal={Journal of Logic Programming},
  year={1997}
}

@article{sebag1996constraint,
  title={{Constraint Inductive Logic Programming}},
  author={Sebag, Mich{\`e}le and Rouveirol, C{\'e}line},
  journal={Advances in ILP},
  pages={277--294},
  year={1996},
  publisher={IOS Press}
}

@article{karalivc1997first,
  title={{First Order Regression}},
  author={Karali{\v{c}}, Aram and Bratko, Ivan},
  journal={Machine learning},
  volume={26},
  number={2},
  pages={147--176},
  year={1997},
  publisher={Springer}
}

@inproceedings{anthony1997generating,
  title={{Generating Numerical Literals During Refinement}},
  author={Anthony, Simon and Frisch, Alan M},
  booktitle={International Conference on Inductive Logic Programming},
  pages={61--76},
  year={1997},
  organization={Springer}
}

@article{kliegr2023qcba,
  title={{QCBA: Improving Rule Classifiers Learned from Quantitative Data by Recovering Information Lost by Discretisation}},
  author={Kliegr, Tom{\'a}{\v{s}} and Izquierdo, Ebroul},
  journal={Applied Intelligence},
  volume={53},
  number={18},
  pages={20797--20827},
  year={2023},
  publisher={Springer}
}

@article{blockeel1998top,
  title={{Top-down Induction of First-order Logical Decision Trees}},
  author={Blockeel, Hendrik and De Raedt, Luc},
  journal={Artificial intelligence},
  volume={101},
  number={1-2},
  pages={285--297},
  year={1998},
  publisher={Elsevier}
}

@inproceedings{cyrus2025inductive,
  title={{An Inductive Logic Programming Approach for Feature-range Discovery}},
  author={Cyrus, Daniel and Varghese, Dany and Tamaddoni-Nezhad, Alireza},
  booktitle={Learning and Reasoning: 4th International Joint Conference on Learning and Reasoning, IJCLR 2024, and 33rd International Conference on Inductive Logic Programming, ILP 2024, Nanjing, China, September 20--22, 2024, Proceedings},
  pages={203},
  year={2025},
  organization={Springer Nature}
}

@article{srinivasan1999numerical,
  title={{Numerical Reasoning with an ILP System Capable of Lazy Evaluation and Customised Search}},
  author={Srinivasan, Ashwin and Camacho, Rui},
  journal={The Journal of Logic Programming},
  volume={40},
  number={2-3},
  pages={185--213},
  year={1999},
  publisher={Elsevier}
}

@inproceedings{belle2020smt+,
  title={{SMT + ILP}},
  author={Belle, Vaishak},
  booktitle={Ninth International Workshop on Statistical Relational AI},
  year={2020}
}

@inproceedings{de2008z3,
  title={{Z3: An Efficient SMT Solver}},
  author={De Moura, Leonardo and Bj{\o}rner, Nikolaj},
  booktitle={International conference on Tools and Algorithms for the Construction and Analysis of Systems},
  pages={337--340},
  year={2008},
  organization={Springer}
}

@inproceedings{varghese2022efficient,
  title={{Efficient abductive learning of microbial interactions using meta inverse entailment}},
  author={Varghese, Dany and Barroso-Bergada, Didac and Bohan, David A and Tamaddoni-Nezhad, Alireza},
  booktitle={International Conference on Inductive Logic Programming},
  pages={127--141},
  year={2022},
  organization={Springer}
}

@phdthesis{morel2023inductive,
  title={{Inductive Logic Programming as Satisfiability Modulo Theories}},
  author={Morel, Rolf},
  year={2023},
  school={University of Oxford}
}

@inbook{barrett2009satisfiability,
title = {{Satisfiability Modulo Theories}},
author = "Clark Barrett and Roberto Sebastiani and Seshia, {Sanjit A.} and Cesare Tinelli",
year = "2009",
language = "English (US)",
isbn = "9781586039295",
series = "Frontiers in Artificial Intelligence and Applications",
publisher = "IOS Press",
number = "1",
pages = "825--885",
booktitle = "Handbook of Satisfiability",
edition = "1",
}

@inproceedings{belle2015probabilistic,
  title={{Probabilistic Inference in Hybrid Domains by Weighted Model Integration}},
  author={Belle, Vaishak and Passerini, Andrea and Van den Broeck, Guy},
  booktitle={Proceedings of the Twenty-Fourth International Joint Conference on Artificial Intelligence, IJCAI 2015, Buenos Aires, Argentina, July 25-31, 2015},
  pages={2770--2776},
  year={2015},
  organization={IJCAI}
}

@inproceedings{chistikov2015approximate,
  title={{Approximate Counting in SMT and Value Estimation for Probabilistic Programs}},
  author={Chistikov, Dmitry and Dimitrova, Rayna and Majumdar, Rupak},
  booktitle={Tools and Algorithms for the Construction and Analysis of Systems: 21st International Conference, TACAS 2015, Held as Part of the European Joint Conferences on Theory and Practice of Software, ETAPS 2015, London, UK, April 11-18, 2015, Proceedings 21},
  pages={320--334},
  year={2015},
  organization={Springer}
}

@inproceedings{de2018learning,
  title={{Learning Constraints from Examples}},
  author={De Raedt, Luc and Passerini, Andrea and Teso, Stefano},
  booktitle={Proceedings of the AAAI conference on artificial intelligence},
  volume={32},
  number={1},
  year={2018}
}

@article{evans2018learning,
  title={{Learning Explanatory Rules from Noisy Data}},
  author={Evans, Richard and Grefenstette, Edward},
  journal={Journal of Artificial Intelligence Research},
  volume={61},
  pages={1--64},
  year={2018}
}

@inproceedings{gao2013dreal,
  title={{dReal: An SMT Solver for Nonlinear Theories Over the Reals}},
  author={Gao, Sicun and Kong, Soonho and Clarke, Edmund M},
  booktitle={International conference on automated deduction},
  pages={208--214},
  year={2013},
  organization={Springer}
}

@article{jaffar1994constraint,
  title={{Constraint Logic Programming: A Survey}},
  author={Jaffar, Joxan and Maher, Michael J},
  journal={The journal of logic programming},
  volume={19},
  pages={503--581},
  year={1994},
  publisher={Elsevier}
}

@inproceedings{kersting2011statistical,
  title={{Statistical Relational AI: Logic, Probability and Computation}},
  author={Kersting, Kristian and Natarajan, Sriraam and Poole, David},
  booktitle={Proceedings of the 11th International Conference on Logic Programming and Nonmonotonic Reasoning (LPNMR’11)},
  pages={1--9},
  year={2011}
}

@inproceedings{kolb2018learning,
  title={{Learning SMT (LRA) Constraints Using SMT Solvers}},
  author={Kolb, Samuel Maria and Teso, Stefano and Passerini, Andrea and De Raedt, Luc and others},
  booktitle={Proceedings of the Twenty-Seventh International Joint Conference on Artificial Intelligence ((IJCAI-18)},
  pages={2333--2340},
  year={2018},
  organization={IJCAI}
}

@inproceedings{martin1997learning,
  title={{Learning Linear Constraints in Inductive Logic Programming}},
  author={Martin, Lionel and Vrain, Christel},
  booktitle={European Conference on Machine Learning},
  pages={162--169},
  year={1997},
  organization={Springer}
}

@inproceedings{mocanu2019pac+,
  title={{PAC + SMT}},
  author={Mocanu, Ionela G and Belle, Vaishak and Juba, Brendan},
  booktitle={Knowledge Representation \& Reasoning Meets Machine Learning: Workshop at NeurIPS'19},
  year={2019}
}

@inproceedings{molina2018mixed,
  title={{Mixed Sum-product Networks: A Deep Architecture for Hybrid Domains}},
  author={Molina, Alejandro and Vergari, Antonio and Di Mauro, Nicola and Natarajan, Sriraam and Esposito, Floriana and Kersting, Kristian},
  booktitle={Proceedings of the AAAI Conference on Artificial Intelligence},
  volume={32},
  number={1},
  year={2018}
}

@article{muggleton1994inductive,
  title={{Inductive Logic Programming: Theory and methods}},
  author={Muggleton, Stephen and De Raedt, Luc},
  journal={The Journal of Logic Programming},
  volume={19},
  pages={629--679},
  year={1994},
  publisher={Elsevier}
}

@article{muggleton1995inverse,
  title={{Inverse Entailment and Progol}},
  author={Muggleton, Stephen},
  journal={New generation computing},
  volume={13},
  pages={245--286},
  year={1995},
  publisher={Springer}
}

@article{muggleton2012ilp,
  title={{ILP Turns 20: Biography and Future Challenges}},
  author={Muggleton, Stephen and De Raedt, Luc and Poole, David and Bratko, Ivan and Flach, Peter and Inoue, Katsumi and Srinivasan, Ashwin},
  journal={Machine learning},
  volume={86},
  pages={3--23},
  year={2012},
  publisher={Springer}
}

@article{cropper2022inductive,
  title={{Inductive Logic Programming at 30}},
  author={Cropper, Andrew and Duman{\v{c}}i{\'c}, Sebastijan and Evans, Richard and Muggleton, Stephen H},
  journal={Machine Learning},
  volume={111},
  number={1},
  pages={147--172},
  year={2022},
  publisher={Springer}
}

@article{muggleton1991inductive,
  title={{Inductive Logic Programming}},
  author={Muggleton, Stephen},
  journal={New generation computing},
  volume={8},
  pages={295--318},
  year={1991},
  publisher={Springer}
}

@inproceedings{shindo21,
  title={{Differentiable Inductive Logic Programming for Structured Examples}},
  author={Shindo, Hikaru and Nishino, Masaaki and Yamamoto, Akihiro},
  booktitle={Proceedings of the AAAI Conference on Artificial Intelligence},
  volume={35},
  number={6},
  pages={5034--5041},
  year={2021}
}

@inproceedings{krishnan21,
  title={{Learning Rules with Stratified Negation in Differentiable ILP}},
  author={Krishnan, Giri P and Maier, Frederick and Ramyaa, Ramyaa},
  booktitle={Advances in Programming Languages and Neurosymbolic Systems Workshop},
  year={2021}
}

@book{Bergadano1995,
author = {Bergadano, Francesco and Gunetti, Daniele},
title = {{Inductive Logic Programming: From Machine Learning to Software Engineering}},
year = {1995},
isbn = {0262023938},
publisher = {MIT Press},
address = {Cambridge, MA, USA}
}

@article{bueff2024learning,
  title={{Learning Explanatory Logical Rules in Non-linear Domains: A Neuro-symbolic Approach}},
  author={Bueff, Andreas and Belle, Vaishak},
  journal={Machine Learning},
  volume={113},
  number={7},
  pages={4579--4614},
  year={2024},
  publisher={Springer}
}

@inproceedings{speichert2019learning,
  title={{Learning Probabilistic Logic Programs Over Continuous Data}},
  author={Speichert, Stefanie and Belle, Vaishak},
  booktitle={International Conference on Inductive Logic Programming},
  pages={129--144},
  year={2019},
  organization={Springer}
}

\appendix

\begin{figure}[t]
\renewcommand{\figurename}{Procedure}
\centering
\scalebox{0.80}{
\begin{minipage}{1.3\linewidth}
\small
\textbf{SMT--ILP Learning Loop.}
Given background knowledge $B_0$, examples $(E^+,E^-)$, and hypothesis language $\mathcal{L}_H$, the system iteratively updates a set of accepted rules $H$ as follows:
\begin{algorithmic}[1]
\State $H \gets \emptyset$, \; $B \gets B_0$
\State $Q_{\text{prev}} \gets 0$, \; $\Delta Q \gets +\infty$, \; $t \gets 0$
\While{$\Delta Q > \theta_{\text{conv}}$ \textbf{and} $t < t_{\max}$}
    \Statex \textbf{(1) Structural hypothesis generation (PyGol)}
    \State $\mathcal{C} \gets \mathrm{GenerateClauses}(B, \mathcal{L}_H)$
    \Comment{Bottom-clause / MIE refinement, optional PI}

    \Statex \textbf{(1a) Arithmetic and range learning (Z3, optional)}
    \If{dataset requires arithmetic constraints}
        \State $\mathcal{C}_{\text{arith}} \gets \mathrm{LearnArithmeticRelations}(E^+, E^-)$
        \Comment{Learn linear constraints $a \cdot x + b \cdot y \leq \theta$ via SMT}
        \State $\mathcal{C} \gets \mathcal{C} \cup \mathcal{C}_{\text{arith}}$
    \EndIf
    \If{dataset requires range constraints}
        \State $\mathcal{C}_{\text{range}} \gets \mathrm{LearnRangeRelations}(E^+, E^-)$
        \Comment{Learn interval constraints $\ell < x < u$ via SMT}
        \State $\mathcal{C} \gets \mathcal{C} \cup \mathcal{C}_{\text{range}}$
    \EndIf

    \Statex \textbf{(2) Numerical parameter instantiation (Z3)}
    \For{$c \in \mathcal{C}$}
        \State Formulate MaxSMT instance $\Phi_c$ with parameters $\theta_c$
        \Comment{Encode coverage of $E^+$ and rejection of $E^-$ as soft/hard constraints}
        \State $\theta^\star_c \gets \mathrm{SolveMaxSMT}(\Phi_c)$
        \State $c^\star \gets c[\theta_c := \theta^\star_c]$
    \EndFor

    \Statex \textbf{(3) Theory-level verification and scoring (Z3)}
    \State $\mathcal{V}_t \gets \emptyset$
    \For{$c^\star \in \{c[\theta_c := \theta^\star_c] : c \in \mathcal{C}\}$}
        \If{$B \land c^\star$ is SAT}
            \State Compute positive coverage 
            $\mathrm{cov}^+(c^\star)$ over $E^+$
            \State Compute negative exclusion 
            $\mathrm{exc}^-(c^\star)$ over $E^-$
            \State Compute precision, recall, F1, support, compression
            \State $\mathrm{score}(c^\star) \gets 
                f(\mathrm{cov}^+, \mathrm{exc}^-, \text{precision}, 
                   \text{recall}, \text{F1}, \text{support}, \text{compression})$
            \If{$\mathrm{score}(c^\star) \ge \theta$}
                \State $\mathcal{V}_t \gets \mathcal{V}_t \cup \{c^\star\}$
            \EndIf
        \EndIf
    \EndFor

    \Statex \textbf{(4) Rule accumulation and quality tracking}
    \State $H \gets H \cup \mathcal{V}_t$
    \State $Q_t \gets \mathrm{Quality}(\mathcal{V}_t)$
    \State $\Delta Q \gets Q_t - Q_{\text{prev}}$
    \State $Q_{\text{prev}} \gets Q_t$
    \State $t \gets t + 1$

    \Statex \textbf{(5) Background refinement (filtered, limited)}
    \If{$t < 3$ \textbf{and} $\exists r \in \mathcal{V}_t : \mathrm{precision}(r) > 0.8$}
        \State $H_{\text{bk}} \gets \{r \in \mathcal{V}_t : \mathrm{precision}(r) > 0.8\}$
        \State $B \gets B \cup H_{\text{bk}}$
        \Comment{Add only high-precision rules in first 3 iterations}
    \EndIf
\EndWhile
\Statex \textbf{(6) Post-processing and final selection}
\State Remove contradictory rules from $H$
\State Remove duplicates and degenerate rules
\State Prioritise: arithmetic $\succ$ PyGol-structured $\succ$ others
\State Filter by precision/recall thresholds
\State $H_{\text{final}} \gets \mathrm{Select}(H)$
\Comment{e.g.\ top-$k$, Pareto frontier, or greedy covering}
\State \Return $H_{\text{final}}$
\end{algorithmic}
\caption{Iterative SMT--ILP learning loop used throughout all experiments. PyGol proposes candidate clause structures; Z3 can additionally learn arithmetic and range constraints directly from data, instantiates numerical parameters via MaxSMT, checks theory satisfiability, computes coverage-based scores, and guides rule selection. Background knowledge is refined only with high-precision rules in early iterations. Convergence is based on quality improvement $\Delta Q$ and a maximum iteration budget $t_{\max}$.}
\label{fig:smt-ilp-loop}
\end{minipage}
}
\end{figure}

\end{document}